\documentclass[sigconf]{acmart}
\usepackage{balance}
\AtBeginDocument{%
  \providecommand\BibTeX{{%
    \normalfont B\kern-0.5em{\scshape i\kern-0.25em b}\kern-0.8em\TeX}}}

\copyrightyear{2020}
\acmYear{2020}
\setcopyright{acmlicensed}
\acmConference[SUMAC '20] {2nd Workshop on Structuring and Understanding of Multimedia heritAge Contents}{October 12, 2020}{Seattle, WA, USA}
\acmBooktitle{2nd Workshop on Structuring and Understanding of Multimedia heritAge Contents (SUMAC '20), October 12, 2020, Seattle, WA, USA}
\acmPrice{15.00}
\acmDOI{10.1145/3423323.3423411}
\acmISBN{978-1-4503-8155-0/20/10}

\begin{document}
 \fancyhead{} 
\title{A Generative Adversarial Approach with Residual Learning for Dust and Scratches Artifacts Removal}

\author{Ionu\c{t} Mironic\u{a}}
\email{mironica@adobe.com}
\affiliation{%
  \institution{Adobe Research Romania}
  \city{Bucharest, Romania}
}

\begin{abstract}
Retouching can significantly elevate the visual appeal of photos, but many casual photographers lack the expertise to operate in a professional manner. One particularly challenging task for old photo retouching remains the removal of dust and scratches artifacts. Traditionally, this task has been completed manually with special image enhancement software and represents a tedious task that requires special know-how of photo editing applications. However, recent research utilizing Generative Adversarial Networks (GANs) has been proven to obtain good results in various automated image enhancement tasks compared to traditional methods. This motivated us to explore the use of GANs in the context of film photo editing. In this paper, we present a GAN based method that is able to remove dust and scratches errors from film scans. Specifically, residual learning is utilized to speed up the training process, as well as boost the denoising performance. An extensive evaluation of our model on a community provided dataset shows that it generalizes remarkably well, not being dependent on any particular type of image. Finally, we significantly outperform the state-of-the-art methods and software applications, providing
superior results.

\end{abstract}



\maketitle


Photographic materials are rather unstable compared to other cultural objects and their degradation is much faster than those of paintings, sculptures or architecture. Dust and scratches on original negatives distinctly appear as light or dark colored artifacts on a scan. These unsightly artifacts have become a major consumer concern during the last years. 
 Dust and scratches~\cite{trumpy2015optical} are two completely different phenomena, but they are often categorized together. "Dust" may refer to any small foreign body that is accidentally lying on the ﬁlm surface that obscures the underlying emulsion image, or conversely, it can have a certain degree of transparency. On the other hand, in the case of a "scratch", the ﬁlm material is missing. They can be superﬁcial or they may penetrate the emulsion layer, completely removing parts of the scanned image. 
 In this moment, the only technology that is able to reduce the effect of dust and scratches are scanners with dedicated hardware that detects the error areas~\cite{tsai2001image}. This hardware is either prohibitively expensive or very slow to use. Without hardware assistance, dust and scratch removal algorithms generally resort to blurring, thereby losing image details. Different from existing software solutions as Photoshop or Lightroom~\cite{Photoshop} that alter the entire image appearance, we present a new alternative for dust and scratch detection that effectively differentiates between defects and image details and it is able to fill the gaps with realistic patches. 
 
 Deep neural networks (DNNs) and GANs~\cite{goodfellow2014generative, Radford2015} have shown very promising results for various image restoration tasks~\cite{Lv2018MBLLEN, hu2018exposure, wei2019single}. However, the design of network architectures remains a major challenge for achieving further improvements. In this paper, we first propose an algorithm that is specifically designed to remove dust and scratches errors from old film images. A novel adversarial learning-based model that can exploit the multi-scale redundancies of natural dust and scratches artifacts is proposed. Experimental results show that the proposed method obtaines better results than any application and algorithm available for this particular problem.

\begin{figure*}[!ht]
\vspace*{0.3in}
    \begin{minipage}[t]{1.0\linewidth}
\begin{center}
    \includegraphics[width=\textwidth]{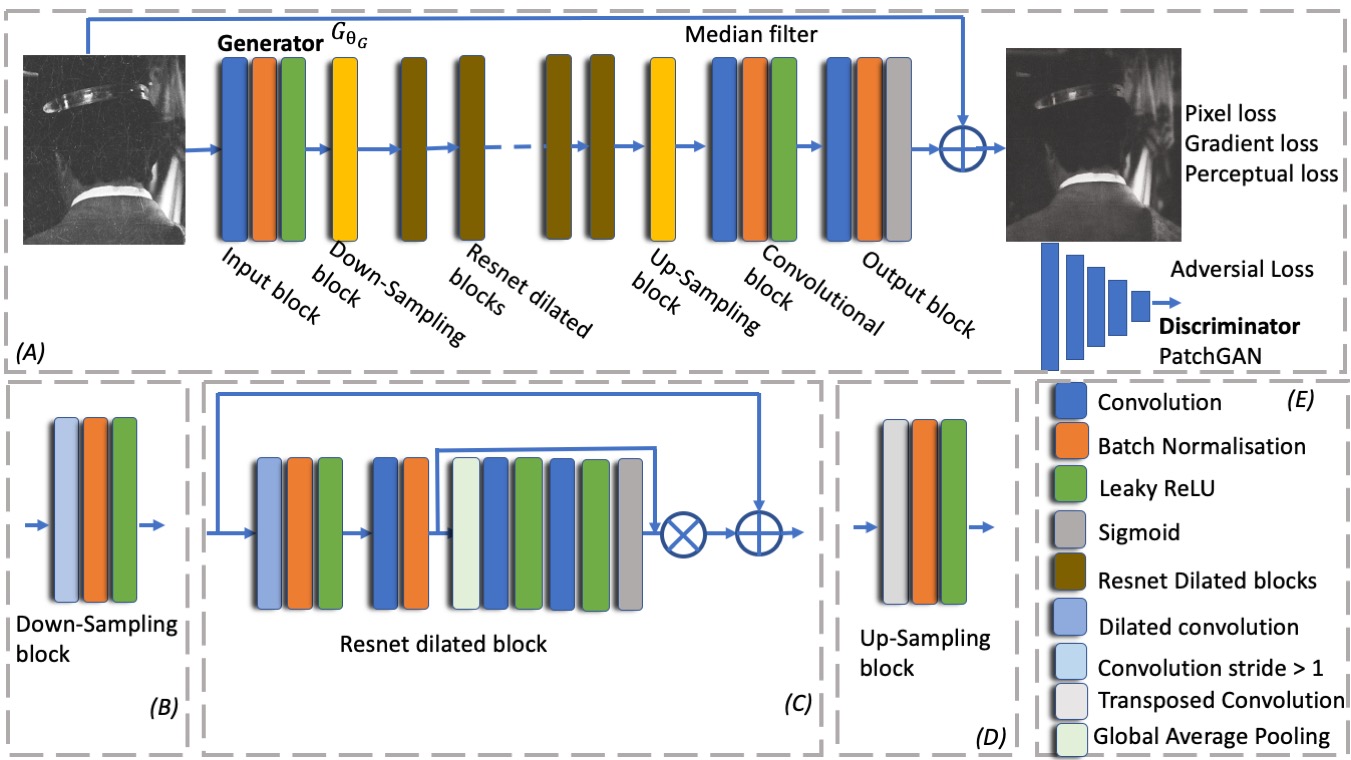}
\end{center}
\caption{Overview of our proposed approach. Figure A presents the overview for the Generator and the Discriminator architectures, while the Figures B, C, D and E expose the technical details of the Generator architecture: the down-sampling and up-sampling blocks, Resnet dilated blocks with the attention mechanism and the agenda with all the used layers on different colors.}\label{figApproach}
\end{minipage}
\end{figure*}

The technical contributions of this work are summarized as follows: 
\begin{itemize}
 
\item Development of a realistic scenario to test the proposed system. In this respect, we create a new benchmark dataset that represents a realistic, natural and challenging scenario in terms of diversity of film photos. The dataset contains pairs of photos with or without artifacts provided by the community.

\item Identify and integrate appropriate architectural features and training procedures which lead to a boosted GAN performance for dust and scratches removal. The proposed steps of improvement include:
\begin{itemize}
\item We introduce a new deep learning representation for the addressed problem of dust and scratches removal model; We find that residual learning can greatly benefit the CNN learning as they can not only speed up the training but also boost the denoising performance;
\item Using a GAN structure, we enable the learning of photo retouching. To our knowledge, this is the first GAN applied to this problem. Furthermore, our approach scales with image resolution and generates no distortion artifacts in the image;

\item We combine several novel loss functions that are able to preserve the visual quality of restored photos.
\item We demonstrate its generality in terms of applications by applying it to a wide category of images;

\item We achieve better performance than current state-of-the-art algorithms and applications.
\end{itemize}
\end{itemize}

The paper has the following structure: Section II reviews related work in the literature, identifying the main drawbacks and possible improvements, Section III details the proposed architecture, Section IV provides implementation details and a quantitative evaluation of the results, while Section V provides conclusions and identifies future work.

\section{Related work}

During the last years, most of the old photo restoration research targeted automatic colorization techniques~\cite{gorriz2019end, nazeri2018image}. Even if the removal of artifacts on films has long been a challenge, not much research has been conducted in this particular field. Most of the approaches involved the breaking down of the problem into two parts: the automatic detection of regions with errors and the in-painting process. In the detection stage the scanned image pixels are categorized using a binary classiﬁcation into two classes: “normal” and “ﬂawed”, thus generating a detection mask~\cite{elgharib2013blotch}. The main problem of this approach is that a low performance of the  detection may generate many unpleasing digital artifacts. A second category of algorithms is inspired by the recent advances of deep learning image segmentation. They are trying to automatically correct the errors without any intermediary process of pixel categorization.

\begin{figure*}[!ht]
\vspace*{0.25in}
    \begin{minipage}[t]{1.0\linewidth}
\begin{center}
    \includegraphics[width=\textwidth]{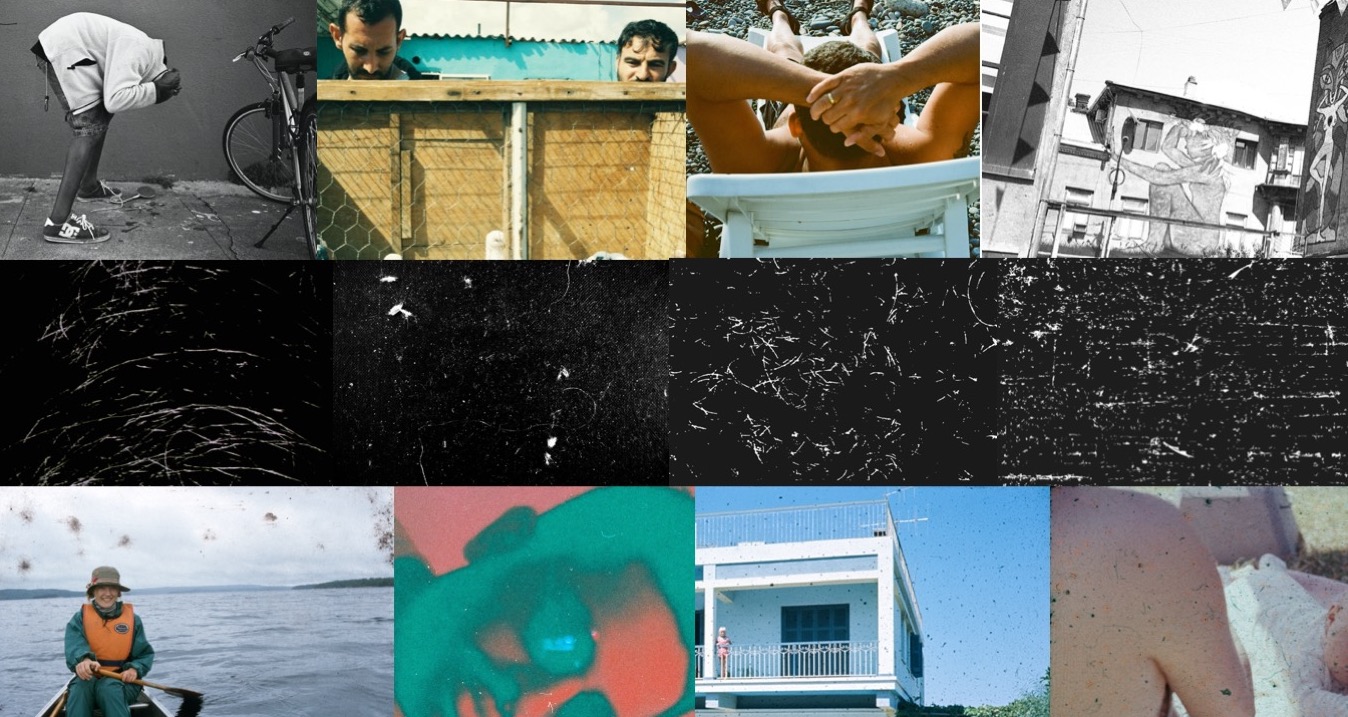}
\end{center}
\caption{Examples of images from the dataset. First row shows training images and the second row provides images with synthetic errors used in the training process and the third row contains images from the test dataset.}\label{figDataset}
\end{minipage}
\end{figure*}

The following in-painting operation is assisted by the detection mask, which indicates where the restoration is necessary and where the image has to be left unedited. In-painting uses the information located near the boundaries of the degraded regions to restore continuous edges by connecting the truncated fragments together. In~\cite{Oliveira2001} an in-painting method based on manual selection of imperfections is introduced. They repair the imperfections by proposing an isotropic diffusion model extended with the notion of user-provided diffusion barriers. A fully automatic solution to detect and remove dust and scratches artifacts is presented in~\cite{bergman2008comprehensive}. They use a local detection step that takes advantage of local contrast difference which enables the algorithm to separate defects from large image edges. Then, they generate a credibility map and apply a weighted bilateral filter for local repairs. An algorithm focused on correction and detection of vertical scratches is presented in~\cite{besserer2004detection}. These types of errors are commonly observed on photographic films due to a mechanical defect in the cameras, but solves only a small part of the scratch errors that may occur. In~\cite{zhou2007removal}, the authors introduced an approach that exploits the image information inside the damaged region. By generating the artifact formation model, they make use of contextual information in the image and creates a color consistency constraint on dust to remove these artifacts. 

Recently, there is an emerging interest in applying deep convolutional neural networks for solving various image-to-image translations problems~\cite{gorriz2019end, errnet, park2018distort}. A similar approach is proposed in~\cite{strubel2019deep}. They adapted a SegNet~\cite{segnet} encoder-decoder architecture trained with cross entropy loss that is able to automatically remove the dust and scratches artifacts. They also proposed a new dataset, but it contains only a limited amount of black and white photo patches. 

However,  the  vast  majority  of  proposed  dust and scratches removal solutions are not able to completely restore the content of the photographs corrupted by noise. Following a careful analysis of  the  existing  methods  and  what  they  are  missing  in  order to create a viable solution, we propose a series of innovative elements that complete the research in this field. Thus, we introduce an appropriate deep learning architecture and training procedures which lead to a boosted GAN performance for dust and scratches removal. We also concern about the size of the proposed model because our aim is  to  develop  a  solution  for  practical  use.  Thus,  we  carefully designed the model so that it has impressive performance using a limited number of learning parameters.
The last, but not the least, we develop a realistic scenario by creating a new diverse and challenging benchmark dataset.
 
\section{Proposed approach}

Inspired by the recent advances in semantic segmentation and GANs, the architecture of the proposed network is illustrated in Figure~\ref{figApproach}. The approach contains two parts: a generator that represents  a feed-forward reconstruction CNN network and a discriminator which is similar to the PatchGAN architecture~\cite{pix2pix, vgg19}. Given an input image $I$ deteriorated by dust and scratches, our goal is to estimate a noise-free image $\hat{T}$. 
Given training image pairs ${((I_n,T_n)}, n = 1, ... ,N$, the following equations should be solved:

\begin{equation}
\hat{\Theta} = argmin_{{\Theta}_G} = \frac{1}{N} \sum\limits_{c=1}^N  {l(G_{{\Theta}_G}(I_n), T_n)}  \label{eq1}
\end{equation}

We first present the details of the generator network $G_{{\Theta}_G}$ followed by the loss functions.

\subsection{Image Reconstruction Network}\label{AA}
Our generator network contains several blocks: input, output, down-sampling, up-sampling and dilated residual blocks. The input block receives a $W\times H\times3$ color image, while the output is bound by a sigmoid function that normalizes the network output values to [0, 1] interval. After the input layer, there is a down-sampling layer that reduces the size of the input. It is followed by seven dilated residual blocks. The architecture of the dilated residual blocks is presented in Figure~\ref{figApproach}C. Each block contains a channel attention mechanism to the feature maps from convolutional layers such that different features are weighted differently according to global statistics of the activations. In order to take information from different scales, the blocks contain dilated convolutions which helps the network to reach a global contextual consistency in the spatial domain.

Each block uses Batch Normalization (BN)~\cite{batchnorm2015}, which has become a common practice in deep learning to accelerate the training of deep neural networks. In the case of Deep Convolutional GANs, it was proven that applying BN to discriminator architectures can be very beneficial to stabilize the GAN learning and to prevent a mode collapse due to poor initialization~\cite{Radford2015}. Internally, BN preserves content-related information by reducing the covariance shift within a mini-batch during training. It uses the internal mean and variance of the batch to normalize each feature channel. After the generator, the discriminator is used in a form of $70 \times 70$ PatchGAN~\cite{pix2pix} fully convolutional architecture. 

\subsection{Residual Learning}\label{RL}
 The discriminator and the generator operate on the residual between the original and a median image. This allows both the generator and discriminator to concentrate only on the important sources of noise and be more specific to the dust and scratches artifacts. Since these regions are challenging to reconstruct and in-painting well, they correspond to the largest perceptual errors. This can also be viewed as subtracting a data dependent baseline which helps to
reduce variance.

\subsection{Training loss}

In our algorithm we want to reduce the dissimilarity between the approximation $\hat{T}$ image and the  correct $T$ image using the Pixel loss, the Gradient Loss and the Perceptual loss. We also want to improve the perceptual quality of the corrected image, for which we have included the Adversarial loss. The perceptual quality of an image $\hat{T}$ represents the degree to which it looks like a natural image and has nothing to do with its similarity to any reference image.

\textbf{Pixel Loss:} We define the Pixel loss between $T$ and $\hat{T}$ images as:
\begin{equation}
L_{Pixel} = ||\hat{T} - T||  \label{eq_pixel_loss}
\end{equation}
where $||.||$ represents the Mean Square Error (MSE) distance.

\textbf{Gradient Loss:} We penalize the difference between edges as the difference of the gradient operator along $x$ and $y$ axes. 
\begin{equation}
L_{Gradient} = ||\nabla_x \hat{T} - \nabla _x T|| + ||\nabla_y \hat{T} - \nabla _y T|| \label{eq_gradient_loss}
\end{equation}
where $\nabla _x$ and $\nabla _y$ represent the gradients over $x$ and $y$ axis.

\textbf{Perceptual Loss:} Metrics such as MSE and SSIM only focus on low-level information in the image, while it is also necessary to use some kind of higher-level information to improve the visual quality. The basic idea is to employ a content extractor. Then, if the enhanced image and the ground truth are similar, their corresponding outputs from the content extractor should also be similar. We use the VGG-19 network~\cite{vgg19} as a content extractor. In particular, we define the context loss based on the output of the ReLU activation layers, by the absolute difference of the representations of the corrected image and the ground truth:
\begin{equation}
L_{Perceptual/i/j}=\frac{1}{W_{i,j}H_{i,j}C_{i,j}} \sum\limits_{x=1}^{W_{i,j}}  \sum\limits_{y=1}^{H_{i,j}} \sum\limits_{z=1}^{C_{i,j}}   || \Phi_{i,j} (\hat{T}) - \Phi_{i,j} (T)||
\end{equation}
where $W_{i,j},H_{i,j},C_{i,j}$ describe the dimensions of the respective feature maps within the VGG-19 network and $\Phi_{i,j}$ represents the feature maps obtained by the $j$-th convolution layer in the $i$-th block. The final $L_{Perceptual}$ become~\cite{Johnson2016Perceptual}:
 \begin{equation}
l_{Perceptual} =  \sum_i || L_{Perceptual/i}(\hat T) - L_{Perceptual/i}(T) || \label{eq8}
\end{equation}
 
 \textbf{Adversarial Loss:}  
 We add an adversarial loss to improve the realism of the corrected images.
 In this way, the generator network $G$ tries to generate film images, while discriminator network $D$ aims to distinguish between fake generated images and real-world images. We define an opponent discriminator $D$ as a VGG \cite{vgg19} architecture while our proposed network acts the role of the generator. We used the adversarial loss function proposed in~\cite{wei2019single}:
\begin{equation}
    L_{Adv} = \mathbb{E}_{x \sim p_x} [\log(D(x))] -\mathbb{E}_{z \sim p_{z}} [\log(1 - D(G(z)))] \label{eq10}
\end{equation}

To summarize, our loss is defined as:
\begin{equation}
L= w_1 L_{Pixel}+ w_2 L_{Gradient}+ w_3 L_{Context}+w_4 L_{Adv}
\end{equation}
where we empirically set the weights as $w_1 = 1, w_2 = 2, w_3 = 1$ and $w_4 = 0.01$ respectively throughout our experiments.

\begin{figure*}[!ht]
\vspace*{0.1in}
\begin{center}
    \includegraphics[width=.9\textwidth]{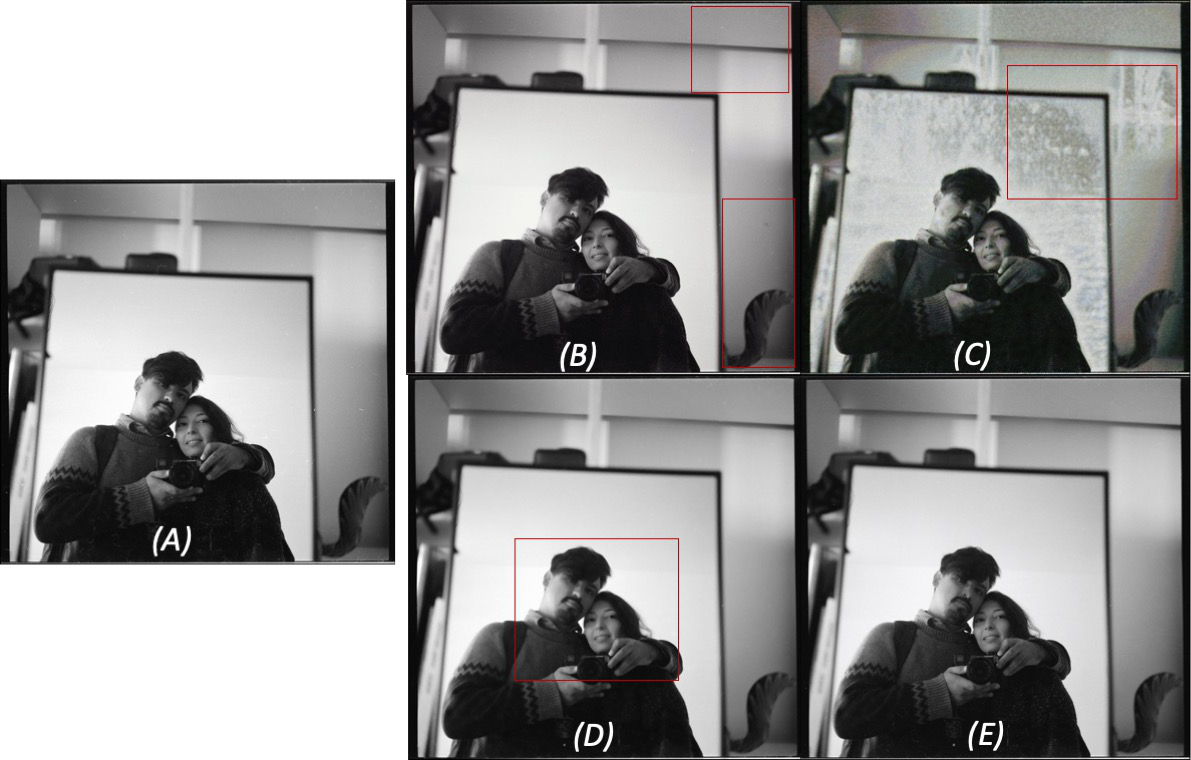}
\end{center}
\caption{Comparison between the proposed approach and other state-of-the-art algorithms. Image (A) represents the input image that contains the dust\&scratches artifacts. Images (B), (C) and (D) provide the output for the~\cite{pix2pix},~\cite{strubel2019deep} and~\cite{Photoshop}. Figure (E) shows the output provided by our approach.}\label{figComparison}
\end{figure*}

\section{Experiments}

\subsection{Implementation Details}
\textbf{Training data.} Real corrected images with ground truth are difficult to obtain. To generate enough training data, we adopt a fusion of synthetic noise and real data as our train dataset. Using a set of $2,500$ film photos without dust and scratches errors, we apply a diverse category of $100$ synthetic artifacts. However, because we want to validate our algorithm on real-world scenarios, for test data we use a set of $500$ photos with real noise that were provided and manually corrected by several professional photographers. Figure~\ref{figDataset} contains several examples from the training dataset, synthetic images and the test dataset provided by the community.

\textbf{Training details.} Our implementation is based on PyTorch. Because we have a reduced number of training photos and their size are higher than $10MP$, we have extracted $128 \times 128$ patches from the training images. We train the model for $200$ epochs using the Adam optimizer~\cite{kingma2014adam}. The base learning rate is set to $5 \times 10^{-4}$ and we decrease the learning rate on each epoch using a standard weight decay that is set to $l/N$, where $l$ represents the learning rate and $N$ is the number of epochs. The weights are initialized as in~\cite{lim2017enhanced}. Because BN usually requires a large batch size value~\cite{errnet}, we set the batch size to a minimum of $32$.

To prevent overfitting, we perform image augmentation, i.e., we apply random transformations to input samples during network training on the fly. We apply random rotation and resizing transformations on both input images and synthetic errors. Also, we apply random intensities of the synthetic images.

\subsection{Experimental Settings}

Through extensive experiments, we qualitatively and quantitatively validate our model and learning framework. We provide full-reference quality assessment using two evaluation parameters: Peak Signal to Noise Ratio (PSNR) and the Structure SIMilarity index (SSIM). PSNR is defined by equation:

\begin{equation}\label{eqn:psnr}
  \mbox{PSNR} = 10\log_{10} \frac{(2^d-1)^2WH}{\sum_{i=1}^W \sum_{j=1}^H (p[i,j]-p'[i,j])^2} 
\end{equation}
where $d$ is the bit depth of pixel, $W$ the image width, $H$ the image height, and $p[i,j]$, 
$p'[i,j]$ is the $i$th-row $j$th-column pixel in the ground truth and corrected image respectively.

The SSIM~\cite{wang2004image} is a method for predicting quality of digital images. is calculated on various windows of an image. The measure between two windows $x$ and $y$ of the same size $N \times N$ is:
\begin{equation}\label{eqn:ssim} 
SSIM(x,y)={\frac {(2\mu _{x}\mu _{y}+c_{1})(2\sigma _{xy}+c_{2})}{(\mu _{x}^{2}+\mu _{y}^{2}+c_{1})(\sigma _{x}^{2}+\sigma _{y}^{2}+c_{2})}}
\end{equation}
where $\mu_x$, $\mu_y$, ${\sigma}_{x}^{2}$ and $\sigma _{y}^{2}$ represents the average and variances of $x$ and $y$, 
$\sigma _{xy}$ the correlation coefficient of x and y and $c_1$ and $c_2$ two variables that stabilize the division with weak denominator.
Larger values of PSNR and SSIM indicate better performance.

\begin{table}[!t]
\vspace*{0.25in}
\caption{Comparison with State-of-the-Art}\label{tblCOmparisonSOA}
\begin{center}
\begin{tabular}{|c|c|c|}
\hline
\textbf{Algorithm} & \textbf{\textit{PSNR}}& \textbf{\textit{SSIM}} \\
\hline
Pix2Pix~\cite{pix2pix}& 31.91& 0.8  \\
\hline
Strubel et al.~\cite{strubel2019deep}& 25.07& 0.65  \\
\hline
Photoshop~\cite{Photoshop} & 22.17& 0.62  \\
\hline
Proposed approach& \textbf{35.12}& \textbf{0.92}  \\
\hline
\end{tabular}
\label{tab1}
\end{center}
\end{table}

\subsection{Comparison with State-of-the-Art}
We compare our method with other deep learning architectures, like GAN-based method Pix2pix-GAN~\cite{pix2pix} and the method proposed in~\cite{strubel2019deep}. For fair comparison, we fine-tune these models on our training dataset. Also, we compare our method with the state-of-the-art algorithm from Adobe Photoshop application~\cite{Photoshop}. Table~\ref{tblCOmparisonSOA} summarizes the results of all competing methods on our test dataset. The quality metrics include PSNR and SSIM.

As can be seen, the proposed model obtains better results than the other methods in terms of PSNR and SSIM index, and the differences in output are evident, as presented in Figure~\ref{figComparison}. The differences between the proposed approach and other algorithm are considerable: Pix2Pix is not able to remove most of the artifacts, the approach proposed in~\cite{strubel2019deep} removes the artifacts but generates color artifacts because it is not suitable for color images. The algorithm embedded in Photoshop is also able to remove most of artifacts but the generated image become blurry and many relevant details are removed from the generated image. Figure~\ref{figComparison}E contains the generated image using the proposed approach that is able to preserve the image details, while removing almost all the dust and scratches artifacts.

To conclude, using the proposed model significantly improves the PSNR and the SSIM, yielding much better results than the state-of-the-art algorithms.

\section{Conclusions}\label{SCM}
The work presented in this paper improved the state-of-the art for automatic dust and scratches removal on film photos using conditional adversarial networks. The proposed GAN architecture integrates techniques from the literature to ensure good training stability and to increase the contribution of the adversarial loss during training, which prevents the GAN from collapsing into blurry images. 

The main contributions of this paper are: i) the design of a method for the restoration of dust and scratches artifacts ii) introduction of a banchmark dataset that create a  realistic  scenario  to  test  the  proposed system and  iii) a thorough comparison with the state-of-the-art deep learning architectures and software applications. \\
Preliminary experiments with the proposed architecture showed good results for the restoration of other artifacts (i.e. noise removal, digital dust). As future work, we plan to investigate its extension to other single and multiple distortions. Also, we want to extend our research to old film videos, which can provide temporal information regarding dust and scratches.

\begin{figure*}[!ht]
\vspace*{-0.0in}
    \begin{minipage}[t]{1.0\linewidth}
\begin{center}
    \includegraphics[width=\textwidth]{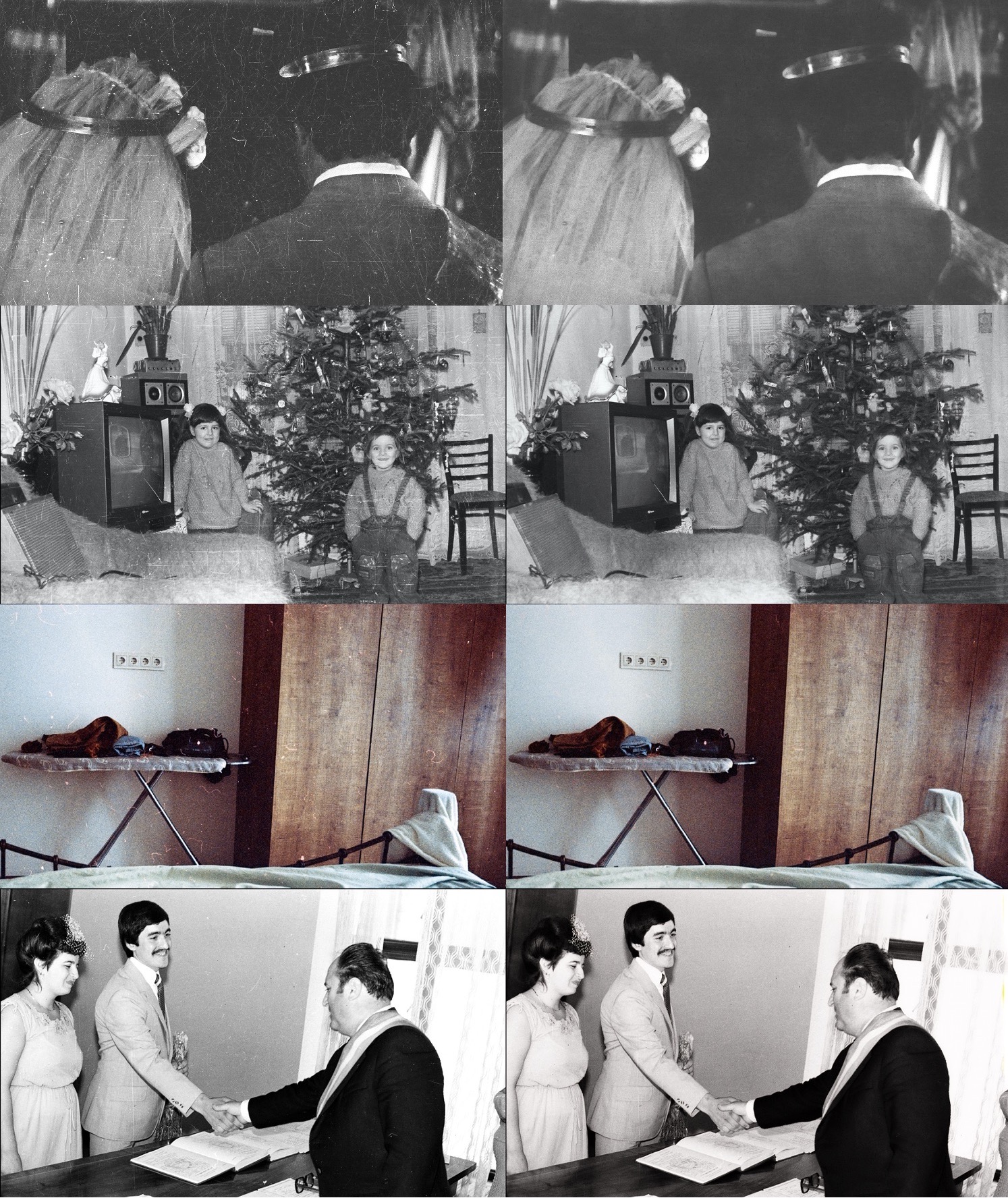}
\end{center}
\caption{Qualitative results of our method. First row shows input images and the second row images provides the output of the proposed algorithm}\label{figExamples1}
\end{minipage}
\end{figure*}

\begin{figure*}[!ht]
\hspace*{-0.0in}
    \begin{minipage}[t]{1.0\linewidth}
\begin{center}
    \includegraphics[width=\textwidth]{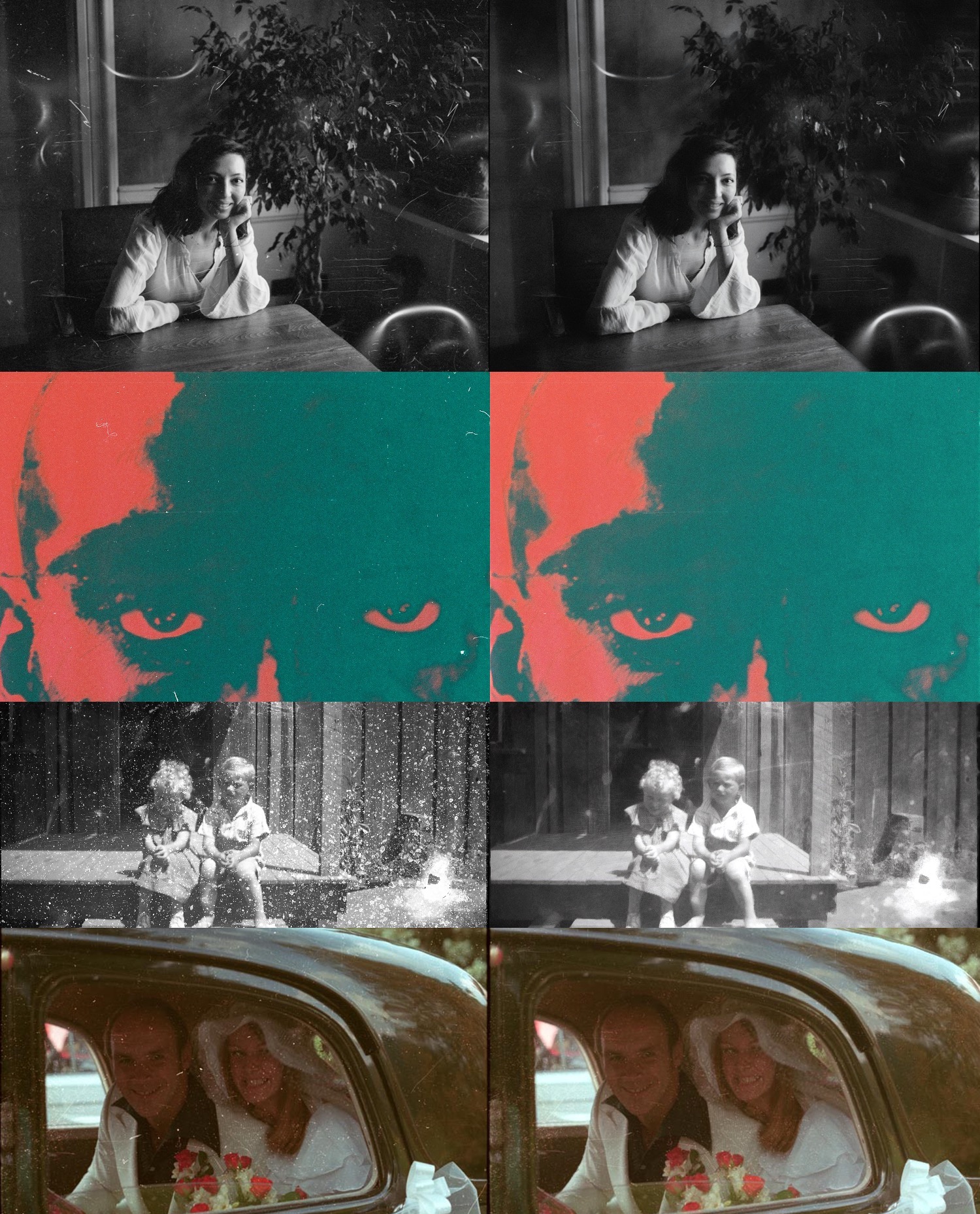}
\end{center}
\end{minipage} 
\end{figure*}
 
\section*{Acknowledgment}
This work would not be possible without the help of the photographers community, especially of Oscar Bola\~{n}os and Mark Sperry and that help us to collect the test dataset. We also express our thanks to all the photographers that contributes with their photos. The last bu not the least, I would like to thanks to Andreea Bîrhal\u{a} and Adrian-Stefan Ungureanu for the interesting discussions held about the topic.
\bibliographystyle{ACM-Reference-Format}

\balance
\bibliography{acmart}

\end{document}